\pdfoutput=1

\documentclass[11pt]{article}

\usepackage{acl}

\usepackage{times}
\usepackage{latexsym}

\usepackage[T1]{fontenc}

\usepackage{enumitem}
\usepackage[group-separator={,}]{siunitx}
\usepackage{multicol, multirow}
\usepackage{graphicx}
\usepackage{tabularx}

\usepackage[utf8]{inputenc}

\usepackage{microtype}

\usepackage{inconsolata}

\usepackage{graphicx}

\title{ToVo: Toxicity Taxonomy via Voting}

\author{
Tinh Son Luong\textsuperscript{\rm 1}\thanks{\ \ Equal contribution.},
Thanh-Thien Le\textsuperscript{\rm 2}\footnotemark[1],
Thang Viet Doan\textsuperscript{\rm 3}\footnotemark[1], \\
\bf{Linh Ngo Van}\textsuperscript{\rm 4},
Thien Huu Nguyen\textsuperscript{\rm 2,5},
Diep Thi-Ngoc Nguyen\textsuperscript{1}\thanks{\ \ Corresponding author.}\\
\textsuperscript{\rm 1}Oraichain Labs \quad
\textsuperscript{\rm 2}VinAI Research \quad
\textsuperscript{\rm 3}Florida International University \\
\textsuperscript{\rm 4}Hanoi University of Science and Technology \quad
\textsuperscript{\rm 5}University of Oregon \\
\texttt{tinh.ls@orai.io}, \quad
\texttt{v.thienlt3@vinai.io}, \quad
\texttt{tdoan011@fiu.edu}, \\
\texttt{linhnv@soict.hust.edu.vn}, \quad
\texttt{thien@cs.oregon.edu}, \quad
\texttt{noah@orai.io}
}

\begin{document}
\maketitle
\begin{abstract}
Existing toxic detection models face significant limitations, such as lack of transparency, customization, and reproducibility. These challenges stem from the closed-source nature of their training data and the paucity of explanations for their evaluation mechanism. To address these issues, we propose a dataset creation mechanism that integrates voting and chain-of-thought processes, producing a high-quality open-source dataset for toxic content detection. Our methodology ensures diverse classification metrics for each sample and includes both classification scores and explanatory reasoning for the classifications.

We utilize the dataset created through our proposed mechanism to train our model, which is then compared against existing widely-used detectors. Our approach not only enhances transparency and customizability but also facilitates better fine-tuning for specific use cases. This work contributes a robust framework for developing toxic content detection models, emphasizing openness and adaptability, thus paving the way for more effective and user-specific content moderation solutions.
\end{abstract}

\begin{table*}[!t]
\centering
\begin{tabular}{llrlr}
\hline
\textbf{Model/Dataset} & \textbf{Metric} & \textbf{Consensus} & \textbf{Metric} & \textbf{Consensus} \\
\hline
\multirow{6}{*}{Llama Guard 2} & Child Sexual Exploitation & 85.484 & Hate & 82.353 \\
& Indiscriminate Weapons & 92.761 & Intellectual Property & 94.840 \\
& Non-Violent Crimes & 83.279 & Privacy & 86.688 \\
& Sex-Related Crimes & 82.353 & Sexual Content & 76.440 \\
& Specialized Advice & 90.271 & Suicide \& Self-Harm & 97.272 \\
& Violent Crimes & 79.630 & \bf Overall & \bf 86.576 \\
\hline
\multirow{6}{*}{OAIM} & harassment & 78,654 & harassment/threatening & 81.422 \\
& hate & 88.090 & hate/threatening & 85.990 \\
& self-harm & 96.677 & self-harm/instructions & 97.928 \\
& self-harm/intent & 96.508 & sexual & 91.233 \\
& sexual/minors & 94.634 & violence & 89.218 \\
& violence/graphic & 90.612 & \textbf{Overall} & \textbf{90.045} \\
\hline
\multirow{4}{*}{Perspective API} & Identity attack & 87.158 & Insult & 75.912 \\
& Profanity & 87.100 & Severe Toxicity & 74.336 \\
& Threat & 90.060 & Toxicity & 72.063 \\
& & & \bf Overall & \bf 81.013 \\
\hline
{BeaverTails} & 14 other metrics & N/A & \bf Overall & \bf{N/A} \\
\hline
\end{tabular}
\caption{\label{tab:dataset_consensus}
42 toxicity metrics available in \textbf{ToVo}. The quantity \emph{Consensus} denotes the percentage of agreement between the original toxicity API/model and the voting outcomes from multiple LLMs. \textbf{\emph{Note:}} The BeaverTails dataset does not come with a model or API, so the consensus rate cannot be calculated.
}
\end{table*}

\section{Introduction}
Detecting toxicity in text generation to ensure safe interactions between human and Large Language Models \cite{radford2019language, brown2020language,hoffmann2022training, chowdhery2023palm, achiam2023gpt, team2023gemini} is a pivotal research challenge.
Despite numerous publications, most contributions have focused on releasing toxicity benchmarking datasets \cite{realtoxic, toxigen, TET}. Regarding detection mechanisms, the research community still relies heavily on partially or fully closed-source models such as Llama Guard \cite{llamaguard} and OpenAI Moderations\footnote{https://platform.openai.com/docs/guides/moderation} (OAIM).
This dependency introduces significant limitations, including a lack of transparency, customization, and reproducibility.

For instance, users cannot fine-tune these models for their own use cases, such as adapting the detection model to address novel forms of toxicity relevant to unique community standards. Additionally, these models often provide no explanation of their evaluation methods, leading to misunderstandings about the model’s performance and making it challenging to control quality. Furthermore, the training data for these models is predominantly closed-source, which hampers efforts to customize detection models or reproduce their results, thereby hindering improvement in this crucial area.

To address these shortcomings, we need to make the following observations. First, to address the lack of interpretability, when the detection model processes a sample, the output must include a classification score and an accompanying explanation for the classification.
Second, to facilitate user-driven fine-tuning, the original model should be diverse, covering a wide range of toxicity criteria from its inception. To meet these desiderata, we develop a toxic detection dataset creation mechanism via voting and chain of thought.

Overall, our contributions are as follows:
\begin{description}[leftmargin=0pt,itemsep=0pt]
\item[a.] We introduce the \textbf{To}xicity Taxonomy \textbf{Vo}ting (\textbf{ToVo}) dataset, a comprehensive resource that categorizes each sample using a diverse selection of metrics from a pool of $42$ derived from four different moderation tools. This extensive coverage ensures that the dataset addresses multiple aspects of toxic content detection. Each classification outcome is generated by a set of open-source models and includes an explanatory rationale, providing valuable insights into the reasoning behind each classification. This dataset is crucial for developing robust and adaptable toxicity detection models.

\item[b.] We leverage the \textbf{ToVo} dataset to develop adaptive taxonomy classification models, capable of operating effectively with both predefined and user-tailored metrics. To demonstrate their efficiency, we benchmark our models against leading moderation tools such as PerspectiveAPI\footnote{https://www.perspectiveapi.com}, OAIM, and Llama Guard 2 on their respective predefined metrics. Additionally, we conduct rigorous Out-of-Domain benchmarking using an evaluation dataset with metrics unrelated to toxicity, showcasing the versatility and robustness of our models.
\end{description}

\section{Dataset Constructions} \label{sec:construction}

\subsection{Dataset Sourcing}
To create the \textbf{ToVo} dataset, we first compile a collection of prompts paired with their respective responses. We begin by extracting prompts from the \texttt{chat-lmsys-1M} \cite{zheng2023lmsyschat1m} dataset, consisting of 1 million conversations between multiple LLMs and their users. Given that this is a general-purpose dataset, many of its sentences do not contain any toxicity. Therefore, we use HateBERT \cite{caselli2020hatebert} to perform a preliminary filtering process, retaining only prompts whose responses exceed a predefined threshold of toxicity probability; this practice has been previously done by \citet{TET}. Subsequently, we randomly select $10{,}000$ prompts from this filtered subset, including $5{,}000$ prompts from single-turn conversations and $5{,}000$ from multi-turn conversations. From these, we obtain responses by prompting open-source models such as Mistral-Instruct \cite{jiang2023mistral} and Zephyr \cite{tunstall2023zephyr}.

\subsection{Classification Label}
\label{sec:construction:class_label}
To establish a gold-labeled taxonomy dataset, we implement a rigorous voting procedure. One of our motivations is to make our dataset similar to the data users might use to fine-tune the model, which can be heterogeneous in terms of the number and type of toxicity metrics in each sample. As a result, we collect a pool of 42 predefined toxicity metrics from Llama Guard 2 - MLCommons \cite{vidgen2024introducing}, OAIM, Perspective API, and BeaverTails \cite{ji2024beavertails}. For each sample in the filtered subset, we randomly select $1$ to $6$ metrics to classify the sample on. Subsequently, three out of six open-source LLMs, which are listed in Appendix \ref{sec:apdx:vote_models} are randomly selected to vote on whether the sample is positive for each of its selected metrics. Criteria for selection include the model's ability to produce sufficiently accurate classifications while avoiding excessively stringent criteria that might prematurely block prompts.

Following this selection, classification results are generated for each chosen model based on the previously selected metrics. To enhance the interpretability of these results, a Chain-of-Thought \cite{wei2022chain} prompting technique is applied during the generation process. This method facilitates a more nuanced and comprehensive understanding of the classification outcomes.

\begin{table*}[t]
\centering
\begin{tabular}{|lr|lr|lr|}
\hline
\textbf{Model/Dataset} & \textbf{Consensus} & \textbf{Model/Dataset} & \textbf{Consensus} & \textbf{Model/Dataset} & \textbf{Consensus} \\
\hline
\multirow{2}{*}{Llama Guard 2} & 91.187 & \multirow{2}{*}{OAIM} & 89.981 & \multirow{2}{*}{Perspective AI} & 77.557 \\
& \emph{92.264} & & \emph{90.625} & & \emph{80.871} \\ 
\hline
\end{tabular}
\caption{\label{tab:predvsapi_short}
The consensus rate between the outputs from our models and the gold labels obtained via the original API/model. \emph{Italicized} values denote results from the model trained with reasoning.
}
\end{table*}

\subsection{Classification Rationale}

As mentioned earlier, three LLMs determine the classification labels for each sample. To select whose rationale would be used as the primary explanation, we engaged in a ranking procedure for each of the six predetermined open-source models, assessing their consensus rates relative to other models. The consensus rate quantifies the level of agreement between the classification outputs of the focal model and the aggregate classifications generated by all selected models.

After gathering these consensus rates, we used the rationale from the model with the highest consensus rate among those that agreed with the majority classification for each sample. This approach ensured that we chose the most consistent and harmonized classification outcome across models, helping to mitigate discrepancies.




\section{Experiments}

\begin{table*}[!t]
\centering
\begin{tabular}{llrr}
\hline
\bf Model/Dataset & \bf Metric & \textbf{Consensus} & \textbf{\emph{Consensus-R}} \\
\hline
\multirow{10}{*}{Out-of-Domain}
& Educational Content & 90.751 & \emph{92.741} \\
& Health and Wellness & 98.057 & \emph{99.297} \\
& Science and Technology & 37.117 & \emph{76.223} \\
& Arts and Culture & 96.516 & \emph{97.527} \\
& Travel and Adventure & 97.245 & \emph{98.087} \\
& Personal Development & 98.590 & \emph{99.167} \\
& Cooking and Recipes & 91.887 & \emph{95.009} \\
& Gardening and Horticulture & 80.304 & \emph{84.589} \\
& Fitness and Exercise & 64.416 & \emph{86.539} \\
& Financial Literacy & 94.662 & \emph{97.312} \\
& \textbf{Overall} & \bf 84.689 & \bf \emph{92.536} \\
\hline
\end{tabular}
\caption{
The consensus rate between the outputs from our models and the gold labels obtained via our voting process on the Out-of-Domain metrics. \emph{Italicized} values denote results from the model trained with reasoning.
}
\label{tab:ood}
\end{table*}

\subsection{Dataset Alignment Evaluation}
We evaluate the alignment of our dataset with other moderation APIs and models, including Llama Guard 2, OpenAI moderation, and the Perspective API. Specifically, for each metric, we measure the \emph{consensus rate}, which is the percentage of agreement between the gold labels obtained via our voting process and the outputs from the original API/model. The results are presented in Table \ref{tab:dataset_consensus}. OAIM shows the highest overall consensus, indicating it is the most aligned with our dataset, while Perspective API has the lowest, suggesting that it might benefit from further alignment with our voting-based gold labels.

Overall, the observed high consensus rates demonstrate a high level of agreement between our gold labels and the outputs from Llama Guard 2 and OAIM, whereas Perspective API shows more variability. This suggests that our voting process produces reasonable and consistent gold labels for toxicity classifications, particularly among metrics with high consensus rates, as predictions should align closely with the reference models rather than being arbitrary. Additionally, this method is scalable, enabling developers to create their own datasets tailored to specific content and metrics.

\subsection{Baseline Model Training \& Evaluation}
\subsubsection{Training configuration}

We trained two baseline models using the pre-trained \texttt{Mistral-Hermes-2-Pro} \footnote{https://huggingface.co/NousResearch/Hermes-2-Pro-Mistral-7B} from NousResearch with 10,000 samples derived from the voting process, utilizing \texttt{transformers} library \cite{wolf2019huggingface} and LoRA \cite{hu2022lora}. One model outputs reasoning for each classification, while the other provides only the classification results. Both model variations—reasoning and non-reasoning—were fine-tuned using LoRA, with a rank of 16 and an $\alpha$ of 16. The training was conducted on a single A100 GPU (40GB memory) with a batch size of 4 and 8 gradient accumulation steps, resulting in a global batch size of 32. The learning rate was set to $1e-4$, and the models were trained over 2 epochs. The prompt templates used for training the reasoning and non-reasoning models are detailed in Appendix \ref{sec:apdx:prompt}

\subsubsection{Toxicity Taxonomy Evaluation}
To evaluate our baseline models, we test them on a set of $2{,}322$ samples with toxicity-related metrics. We compare the alignment of our baseline models with other moderation tools. The results are presented in Table \ref{tab:predvsapi_short} and Appendix \ref{sec:apdx:moreresult}. It is noteworthy that lower consensus rates in certain metrics when comparing our models with other tools (e.g., "Sexual Content" for Llama Guard 2 and "Toxicity" for Perspective API) do not necessarily suggest areas where our classification criteria might need refinement. Instead, they could indicate that while the metrics are the same, the level of toxicity tolerance that determines whether a sample should be labeled as positive differs between our model and the original API. We also benchmark our models with the labels obtained from the voting process; the details are presented in Appendix \ref{sec:apdx:moreresult}.

\subsubsection{Out-of-Domain Taxonomy Evaluation}
Additionally, we construct an Out-of-Domain (OOD) test set using metrics unrelated to toxicity. This experiment aims to evaluate the generalization of the baseline models and their zero-shot performance on user-custom metrics.

In the OOD test set, there are 10 metrics that are unrelated to the toxicity metrics used in the training set. Using the specific definitions of these metrics, which are detailed in Appendix \ref{sec:apdx:ood_testset}, we apply the construction process described in Section \ref{sec:construction} to create a test set comprising $1741$ samples. To evaluate our models, we compare the consensus rate between our models and the gold labels according to the voting process described in Section \ref{sec:construction}. The result are available in Table \ref{tab:ood}.

Both models achieve high consensus rates for several OOD metrics, especially the reasoning model: it outperforms its non-reasoning counterpart across all metrics. Additionally, some metrics exhibit more variability in consensus rates between the standard and reasoning models. The "Fitness and Exercise" metric, for example, shows a notable improvement from $64.416\%$ to $86.539\%$ with the reasoning model, suggesting that certain categories benefit significantly from the additional interpretability provided by reasoning. The overall consensus rate across all OOD metrics is $84.689\%$ for the non-reasoning model and $92.536\%$ for the reasoning model. This overall improvement underscores the value of the reasoning approach in achieving more reliable and consistent classification outcomes for high-novelty metrics.

This experiment demonstrates that our models, particularly those incorporating reasoning, are highly adaptable and effective in classifying content across a wide range of domains. This adaptability is crucial for real-world applications, enhancing the reliability of the models in handling diverse and dynamic content types.

\section{Conclusions}
In this paper, to address significant limitations in existing toxic content detection models, we introduce the \textbf{To}xicity Taxonomy \textbf{Vo}ting (\textbf{ToVo}) dataset, developed through a rigorous voting mechanism and Chain-of-Thought prompting to ensure high-quality, explainable classification outcomes.

Utilizing \textbf{ToVo}, we train two taxonomy models that perform exceptionally well on toxicity-related metrics in the evaluation dataset. These models not only align closely with the gold labels generated by the voting process but also demonstrate a high level of consensus with other moderation tools. Furthermore, our reasoning model achieves exceptional results on the Out-of-Domain test set, affirming its adaptability to user-specific custom metrics. This underscores the model's potential for fine-tuning in diverse application scenarios.

Our work introduces a novel method for creating customized moderation tools effortlessly, using an automated process that combines Voting and Chain-of-Thought techniques. By fostering safer human-LLM interactions and empowering users with customizable moderation tools, we hope this work can pave the way for creating a safer and more inclusive digital environment.



\section*{Limitations \& Future Works}
While the voting process with Chain-of-Thought prompting is efficient, generating large amounts of data remains time-consuming and labor-intensive. Additionally, utilizing large language models (LLMs) for taxonomy purposes results in slow inference speeds. This slowdown is attributed to the substantial size of the models (7 billion parameters) and the inclusion of reasoning in some responses, which further hampers processing speed.

Moreover, the current version of our taxonomy only supports binary classification. This binary approach can sometimes be insufficient, as classifying metrics solely as 0 or 1 may not accurately capture the nuances of their potential unsafety.

In the future, we aim to extend our research beyond large language models (LLMs) to widen its range of applications. Specifically, we plan to apply our toxicity detection framework to human-human interactions, web content, and online forums. By doing so, we must enhance the robustness and versatility of our models, ensuring they can effectively handle diverse and dynamic contexts of toxic content across various platforms. This expansion will also involve refining our taxonomy to support more nuanced classifications, enabling more accurate and context-sensitive moderation. Ultimately, our goal is to contribute to safer and more inclusive digital environments through adaptable toxicity detection solutions.


\bibliography{anthology,custom}

\appendix

\section{Appendix}
\label{sec:appendix}
\subsection{Voting Models}
\label{sec:apdx:vote_models}
The six LLMs we use during our Voting Process described in Section \ref{sec:construction:class_label} are:
\begin{itemize}
    \item \texttt{deepseek-llm-67b-chat} \cite{deepseek-llm},
    \item \texttt{dolphin-2.5-mixtral-8x7bv}\footnote{https://huggingface.co/cognitivecomputations/dolphin-2.5-mixtral-8x7b},
    \item \texttt{Nous-Hermes-2-Mixtral-8x7B-DPO} \cite{Nous-Hermes-2-Mixtral-8x7B-DPO},
    \item \texttt{WizardLM-2-8x22B}\footnote{https://huggingface.co/alpindale/WizardLM-2-8x22B},
    \item \texttt{Yi-34B-Chat} \cite{ai2024yi}, and
    \item \texttt{Mixtral-8x7B-Instruct-v0.1} \cite{jiang2024mixtral}.
\end{itemize}

\subsection{Out-of-Domain Test Set}
\label{sec:apdx:ood_testset}
In the Out-of-Domain Test Set, there are 10 metrics that are unrelated to the toxicity metrics used in the training set. The specific descriptions of these 10 metrics are presented in Table \ref{tab:ood_explain}. Using these definitions, we apply the construction process described in Section \ref{sec:construction} to create a test set comprising $1741$ samples.

\subsection{Prompt Templates}
\label{sec:apdx:prompt}

\begin{figure*}
    \centering
    \includegraphics{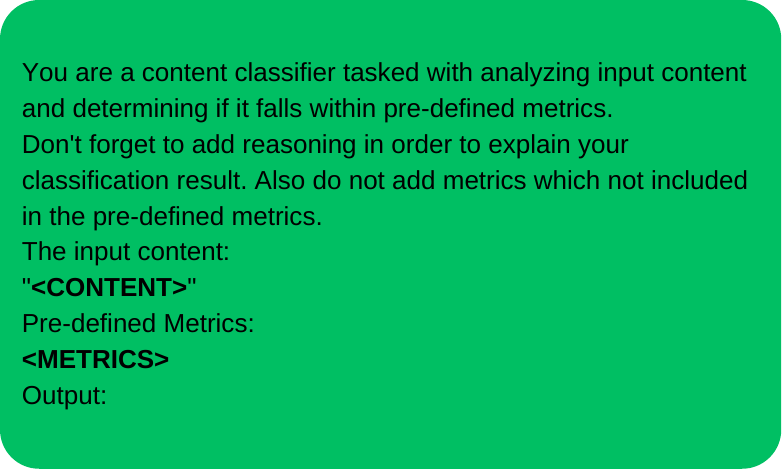}
    \caption{Prompt template for the training of the reasoning model.}
    \label{fig:promptr}
\end{figure*}

\begin{figure*}
    \centering
    \includegraphics{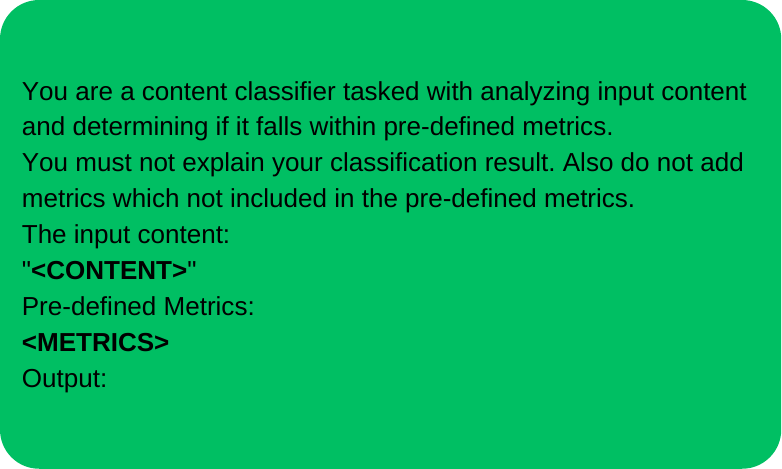}
    \caption{Prompt template for the training of the non-reasoning model.}
    \label{fig:promptnr}
\end{figure*}

The prompt templates used for training the reasoning and non-reasoning models are detailed in Figures \ref{fig:promptr} and \ref{fig:promptnr}, respectively.

\subsection{Additional Experimental Results}
\label{sec:apdx:moreresult}

Table \ref{tab:predvsapi_lg2}, \ref{tab:predvsapi_oaim}, and \ref{tab:predvsapi_papi} present the performance of our models, using the gold labels from the original toxicity detection API/model.

Table \ref{tab:pred_voting_llamaguard2}, \ref{tab:pred_voting_oaim}, \ref{tab:pred_voting_papi}, and \ref{tab:pred_voting_beaver} present the performance of our models, using the gold labels from our voting process.





\subsection{Experiments on BeaverTails-30K test set}
To further evaluate our models against others such as MD Judge and Llama Guard 2, we conduct experiments using the BeaverTails-30K test set. It is important to note that while our model is trained on a dataset incorporating metrics similar to those used in BeaverTails, it is not trained directly on the BeaverTails dataset.

The results of our experiments are presented in Table \ref{tab:beavertail30kres}. From here we can make the following observations: 
\begin{itemize}
\item Solely relying on the metric's description to classify content as safe or unsafe is insufficient. It is crucial to also consider the degree to which the content may be deemed unsafe.
\item For instance, Some prompts in BeaverTails involve "how to prank..." scenarios. While most of these are classified as Non-Violence Unethical Behaviors, which is safe, both our models and the voting results from other models suggest that such content has the potential to be unsafe. However, without specific details about the prank, it is challenging to classify the content as unsafe within the BeaverTails framework. This suggests that the metric descriptions in BeaverTails may not fully capture the nuances of their manual labeling process.
\item We attempted to broaden the scope of the metric descriptions to better classify potential cases as unsafe. This adjustment resulted in a slight improvement in our model's performance on the BeaverTails-30K test set. Consequently, we believe that obtaining metric descriptions that more closely align with BeaverTails' labeling criteria could significantly enhance the accuracy of our models or others.
\end{itemize}









\begin{table*}
\centering
\begin{tabularx}{\textwidth}{|l|X|}
\hline
Metric & Description \\
\hline
Educational Content &
This includes content designed to educate and inform users on a variety of topics, such as academic subjects, professional skills, hobbies, and general knowledge. It aims to provide learning opportunities and improve understanding. \\
        \hline
        Health and Wellness & This refers to content that promotes physical, mental, and emotional well-being. It includes fitness tips, nutritional advice, mental health resources, and general wellness information. \\
        \hline
        Science and Technology & This category encompasses content related to scientific discoveries, technological advancements, and innovations. It includes research findings, tech reviews, and discussions on scientific topics. \\
        \hline
        Arts and Culture & This includes content that explores various forms of art and cultural expressions. It covers topics such as visual arts, music, literature, theater, and cultural traditions from around the world. \\
        \hline
        Travel and Adventure & This pertains to content that inspires and informs about travel destinations, experiences, and adventures. It includes travel guides, adventure stories, and tips for travelers. \\
        \hline
        Personal Development & This category includes content aimed at personal growth and self-improvement. It covers topics such as goal setting, productivity, motivation, and skills development. \\
        \hline
        Cooking and Recipes & This includes content related to culinary arts, recipes, cooking tips, and food preparation techniques. It aims to inspire and guide individuals in creating delicious meals. \\
        \hline
        Gardening and Horticulture & This pertains to content about gardening practices, plant care, landscaping, and horticultural techniques. It includes advice on growing flowers, vegetables, and maintaining gardens. \\
        \hline
        Fitness and Exercise & This includes content focused on physical fitness, workout routines, exercise techniques, and sports. It aims to promote an active lifestyle and provide guidance on maintaining physical health. \\
        \hline
        Financial Literacy & This category includes content that educates individuals on managing finances, budgeting, investing, and achieving financial goals. It aims to improve financial understanding and decision-making. \\
        \hline
\end{tabularx}
\caption{\label{tab:ood_explain}
Descriptions of the Out-of-Domain metrics.
}
\end{table*}

\begin{table*}[!t]
\centering
\begin{tabular}{llrr}
\hline
\bf Model/Dataset & \bf Metric & \textbf{Consensus} & \bf \emph{Consensus-R} \\
\hline
\multirow{12}{*}{Llama Guard 2}
& Child Sexual Exploitation & 90.659 & \emph{90.643} \\
& Indiscriminate Weapons & 97.959 & \emph{97.850} \\
& Non-Violent Crimes & 89.474 & \emph{90.798} \\
& Sex-Related Crimes & 82.486 & \emph{87.500} \\
& Specialized Advice & 96.809 & \emph{97.802} \\
& Violent Crimes & 89.286 & \emph{91.926} \\
& Hate & 86.700 & \emph{87.113} \\
& Intellectual Property & 99.471 & \emph{96.757} \\
& Privacy & 93.252 & \emph{96.795} \\
& Sexual Content & 77.515 & \emph{77.640} \\
& Suicide \& Self-Harm & 97.382 & \emph{98.387} \\
& Overall & 91.187 & \emph{92.264} \\
\hline
\end{tabular}
\caption{\label{tab:predvsapi_lg2}
The consensus rate between the outputs from our models and the gold labels obtained via the original API/model on the metrics from Llama Guard 2. \emph{Italicized} values denote results from the model trained with reasoning.
}
\end{table*}

\begin{table*}[!t]
\centering
\begin{tabular}{llrr}
\hline
\bf Model/Dataset & \bf Metric & \textbf{Consensus} & \bf \emph{Consensus-R} \\
\hline
\multirow{12}{*}{OAIM}
& Harassment & 73.714 & \emph{78.049} \\
& Hate & 88.095 & \emph{87.180} \\
& Self-Harm & 98.305 & \emph{97.647} \\
& Self-Harm/Intent & 100.000 & \emph{99.492} \\
& Sexual/Minors & 93.367 & \emph{95.676} \\
& Violence/Graphic & 89.560 & \emph{91.954} \\
& Harassment/Threatening & 86.473 & \emph{85.714} \\
& Hate/Threatening & 84.536 & \emph{84.946} \\
& Self-Harm/Instructions & 95.238 & \emph{94.479} \\
& Sexual & 87.817 & \emph{90.426} \\
& Violence & 91.848 & \emph{90.173} \\
& Overall & 89.981 & \emph{90.625} \\
\hline
\end{tabular}
\caption{\label{tab:predvsapi_oaim}
The consensus rate between the outputs from our models and the gold labels obtained via the original API/model on the metrics from OpenAI Moderation. \emph{Italicized} values denote results from the model trained with reasoning.
}
\end{table*}

\begin{table*}[!t]
\centering
\begin{tabular}{llrr}
\hline
\bf Model/Dataset & \bf Metric & \textbf{Consensus} & \bf \emph{Consensus-R} \\
\hline
\multirow{7}{*}{Perspective AI}
& Identity attack & 91.667 & \emph{93.902} \\
& Profanity & 84.211 & \emph{86.264} \\
& Threat & 89.063 & \emph{90.000} \\
& Insult & 68.023 & \emph{73.913} \\
& Severe Toxicity & 67.172 & \emph{72.251} \\
& Toxicity & 65.946 & \emph{69.663} \\
& Overall & 77.557 & \emph{80.871} \\
\hline
\end{tabular}
\caption{\label{tab:predvsapi_papi}
The consensus rate between the outputs from our models and the gold labels obtained via the original API/model on the metrics from Perspective API. \emph{Italicized} values denote results from the model trained with reasoning.
}
\end{table*}

\begin{table*}[!t]
\centering
\begin{tabular}{llrr}
\hline
\bf Model/Dataset & \bf Metric & \textbf{Consensus} & \bf \emph{Consensus-R} \\
\hline
\multirow{12}{*}{Llama Guard 2}
& Child Sexual Exploitation & 96.133 & \emph{97.647} \\
& Indiscriminate Weapons & 100.000 & \emph{99.462} \\
& Non-Violent Crimes & 89.412 & \emph{90.124} \\
& Sex-Related Crimes & 93.220 & \emph{90.533} \\
& Specialized Advice & 97.340 & \emph{96.721} \\
& Violent Crimes & 94.048 & \emph{95.652} \\
& Hate & 94.555 & \emph{92.228} \\
& Intellectual Property & 99.471 & \emph{99.460} \\
& Privacy & 93.827 & \emph{93.548} \\
& Sexual Content & 97.024 & \emph{98.125} \\
& Suicide \& Self-Harm & 97.906 & \emph{97.861} \\
& Overall & 95.833 & \emph{95.657} \\
\hline
\end{tabular}
\caption{\label{tab:pred_voting_llamaguard2}
The consensus rate between the outputs from our models and the gold labels obtained via our voting process on the metrics from Llama Guard 2. \emph{Italicized} values denote results from the model trained with reasoning.
}
\end{table*}

\begin{table*}[!t]
\centering
\begin{tabular}{llrr}
\hline
\bf Model/Dataset & \bf Metric & \textbf{Consensus} & \bf \emph{Consensus-R} \\
\hline
\multirow{12}{*}{OAIM}
& Harassment & 90.857 & \emph{89.091} \\
& Hate & 88.691 & \emph{91.667} \\
& Self-Harm & 98.864 & \emph{99.412} \\
& Self-Harm/Intent & 99.034 & \emph{98.469} \\
& Sexual/Minors & 96.939 & \emph{96.216} \\
& Violence/Graphic & 93.407 & \emph{97.126} \\
& Harassment/Threatening & 87.864 & \emph{88.205} \\
& Hate/Threatening & 86.598 & \emph{88.172} \\
& Self-Harm/Instructions & 99.405 & \emph{99.387} \\
& Sexual & 97.449 & \emph{96.791} \\
& Violence & 93.478 & \emph{93.103} \\
& Overall & 93.860 & \emph{94.311} \\
\hline
\end{tabular}
\caption{\label{tab:pred_voting_oaim}
The consensus rate between the outputs from our models and the gold labels obtained via our voting process on the metrics from Open AI Moderation. \emph{Italicized} values denote results from the model trained with reasoning.
}
\end{table*}

\begin{table*}[!t]
\centering
\begin{tabular}{llrr}
\hline
\bf Model/Dataset & \bf Metric & \textbf{Consensus} & \bf \emph{Consensus-R} \\
\hline
\multirow{7}{*}{Perspective AI}
& Identity attack & 93.413 & \emph{93.865} \\
& Profanity & 92.105 & \emph{94.506} \\
& Threat & 92.188 & \emph{92.265} \\
& Insult & 83.140 & \emph{83.333} \\
& Severe Toxicity & 87.374 & \emph{87.958} \\
& Toxicity & 85.946 & \emph{84.358} \\
& Overall & 89.040 & \emph{89.414} \\
\hline
\end{tabular}
\caption{\label{tab:pred_voting_papi}
The consensus rate between the outputs from our models and the gold labels obtained via our voting process on the metrics from Perspective API. \emph{Italicized} values denote results from the model trained with reasoning.
}
\label{tab
}
\end{table*}

\begin{table*}[t]
\centering
\begin{tabular}{llrr}
\hline
\bf Model/Dataset & \bf Metric & \textbf{Consensus} & \bf \emph{Consensus-R} \\
\hline
\multirow{14}{*}{BeaverTails}
& Animal Abuse & 99.490 & \emph{99.465} \\
& Child Abuse & 97.727 & \emph{96.450} \\
& Controversial Topics, Politics & 85.567 & \emph{86.631} \\
& Discrimination, Stereotype, Injustice & 89.618 & \emph{88.571} \\
& Drug Abuse, Weapons, Banned Substance & 99.320 & \emph{99.254} \\
& Financial Crime, Property Crime, Theft & 98.750 & \emph{100.000} \\
& Hate Speech, Offensive Language & 89.130 & \emph{85.876} \\
& Misinformation Regarding Ethics, Laws, and Safety & 93.258 & \emph{95.210} \\
& Non-Violent Unethical Behavior & 85.807 & \emph{81.879} \\
& Privacy Violation & 95.313 & \emph{94.022} \\
& Self-Harm & 98.438 & \emph{98.387} \\
& Sexually Explicit, Adult Content & 96.354 & \emph{97.778} \\
& Terrorism, Organized Crime & 99.435 & \emph{99.408} \\
& Violence, Aiding and Abetting, Incitement & 91.038 & \emph{91.584} \\
\hline
\end{tabular}
\caption{\label{tab:pred_voting_beaver}
The consensus rate between the outputs from our models and the gold labels obtained via our voting process on the metrics from BeaverTails. \emph{Italicized} values denote results from the model trained with reasoning.
}
\end{table*}

\begin{table*}[t]
    \centering
    \begin{tabular}{ccc}
        \hline
         Model &  Metrics & Unsafe F1 \\
         \hline
         Reasoning & Default BeaverTails Metrics & 0.4104\\ 
         Reasoning & Custom BeaverTails Metrics & 0.4164\\
         \hline
         Voting (200 samples) & Default BeaverTails Metrics & 0.4131\\
         \hline
    \end{tabular}
    \caption{\label{tab:beavertail30kres}
    Performance of Voting process and our reasoning model on BeaverTails-30K test set
    }
\end{table*}

\end{document}